# SARS-CoV-2 virus RNA sequence classification and geographical analysis with convolutional neural networks approach.

Selçuk YAZAR
Trakya University Edirne Vocational College of Technical Sciences

**Abstract**
Covid-19 infection, which spread to the whole world in December 2019 and is still active, caused more than 250 thousand deaths in the world today. Researches on this subject have been focused on analyzing the genetic structure of the virus, developing vaccines, the course of the disease, and its source. In this study, RNA sequences belonging to the SARS-CoV-2 virus are transformed into gene motifs with two basic image processing algorithms and classified with the convolutional neural network (CNN) models. The CNN models achieved an average of 98% Area Under Curve(AUC) value was achieved in RNA sequences classified as Asia, Europe, America, and Oceania. The resulting artificial neural network model was used for phylogenetic analysis of the variant of the virus isolated in Turkey. The classification results reached were compared with gene alignment values in the GISAID database, where SARS-CoV-2 virus records are kept all over the world. Our experimental results have revealed that now the detection of the geographic distribution of the virus with the CNN models might serve as an efficient method.

**Introduction**

Artificial intelligence practices and particularly deep learning studies are a widely used discipline in many research fields, including medicine and bioinformatics. The CNN models, especially in the field of medical imaging, are very successful in lesions and disease diagnosis. In addition to the success of deep learning methods in the fields of image processing, natural language processing, also has a lot of usage on a time scale with approaches such as Long-Short Term memory. In deep learning practices, low-level features such as DNA sequence, pathology images, and tomography scans can be learned from the data, by largely eliminating the need for engineering applications. Another important feature of the deep learning technique is that the intermediate values and weights obtained during training can be used in other related applications.

Deep learning approaches are frequently used in the processing of large-scale genetic data, especially in the field of bioinformatics. Today, the size of these datasets exceeds the size of 10 million [1]. Studies in this area are concentrated in the analysis and classification of DNA and RNA sequences. By using certain machine learning algorithms such as Support Vector Machines, Random Trees, it is aimed to reveal certain or undetermined structures in the genetic sequence with the developed methods. Apart from the classical approaches, the proposed deep learning methods can be used in areas such as metagenomics and proteomics (protein researches) as it has been shown. In these studies, deep learning models were used to analyze splice junction information and gene structures of RNA of viruses and to determine variants. For example, deep learning model variables allow a secondary RNA structure which has been detected, can be used to detect another microRNA insertion lasso [2].

There are several publications about phylogenetic RNA–seq analysis using distinct deep learning algorithms. Applications that investigate the patristic distance (distance from the ancestor) of the RNA sequences obtained and differentiated from various virus variants and the structure of the tree structures are popular. In the CNN methods used in this type of paper here, the MultiDimensional Scaling approach was used and classical methods such as SVM were compared [3]. Another study focuses on the classification of insertion points in RNA-seq matches. In the mentioned paper Homo sapiens Splice Sites Database (HS3D) has been used to classify and variant estimates on mRNA (messenger RNA) using CNN models and compared with other common LSTM and SVM algorithms [4]. In another study, Single-cell RNA-seq analysis used the Encode-Decoder approach, which is utilized for today to reduce noise in deep learning applications [5]. The encoder-decoder approach used in deep learning applications is based on the coding strategy that the network can process before sending data to the artificial neural network. In Encoder-Decoder applications, input data is reduced in the bottleneck layer and then reproduced for further processing. Another area where such studies are concentrated in cancer research [6]. Auto Encoder method was used to classify cancer-causing genes.

The SARS-CoV-2 [7] outbreak poses the biggest global health and socioeconomic threat since the Second World War. Phylogenetic studies have gained increasing importance with regard to the Covid-19 virus, which is currently watching horizontally but is concerned about the epidemic again in the coming days. The geographical spread of the virus and the monitoring of the mutations it undergoes also constitute the basis of the vaccine studies to be produced against the virus. The creation of phylogenetic trees of viruses and bacteria plays a key role in understanding the structures of these pathogens.

In this study, we took the RNA data of the SARS-Cov-2 virus from the databases that published it internationally for researchers to use. We have collected this data in four groups: Asia, Europe, America, and Australia. In the data obtained, we used RNA sequences with an average length of more than 29,000 characters. Then, we aimed to classify the sequences by using the model created based on the DenseNet-121 artificial neural network, using the midpoint circle and Smallest Univalue Segment Assimilating Nucleus (SUSAN) [8] algorithms. We compared the classification results from the models we created with gene alignment counts from viruses isolated in Turkey, Spain, Italy, Iran, and India.

**Material and methods**

**Dataset**

The data of the RNA sequences we used in the study were provided from two databases. Since it was isolated in December 2019, GISAID [9] and the National Center for Biotechnology Center (NCBI Virus) [10] databases provide analysis of SARS-CoV-2 RNA records worldwide. During the study, we obtained the Asian-tagged series from the NCBI Virus database and the others via GISAID. The data stored and published in these repositories are kept in FASTA [11] file format. In this file format, the date, location, quality, and publication information of the researchers are available when the virus's RNA sequence is isolated. Since the quality of RNA sequences is also mentioned in these repositories, we did not prefer the sequences that are incomplete or have low safety in the classification model we created. Today, more than 16 thousand complete or partial genomic sequences are published on GISAID. In the NCBI (Virus) database, more than 2 thousand sequences are published complete or partially tagged. The total number of complete gene sequences obtained from these two databases is 3754. The distribution of the numbers for the examples here is Europe 1402, America 1226, Asia 858 and Australia 268. The distribution of the data is shown in Figure 1. The quantities that determine the quality and completeness of the data in these repositories are that the base pair number is more than 29,000 and the unsolved amino acid value is less than 5%. These features, which we have listed here, have created the motifs of the RNA information that we have collected so that all of the amino acid values are displayed correctly. Although we tried to select fully isolated gene sequences, in some cases we inevitably had to use sequences containing spaces as "N" due to the low number of data.

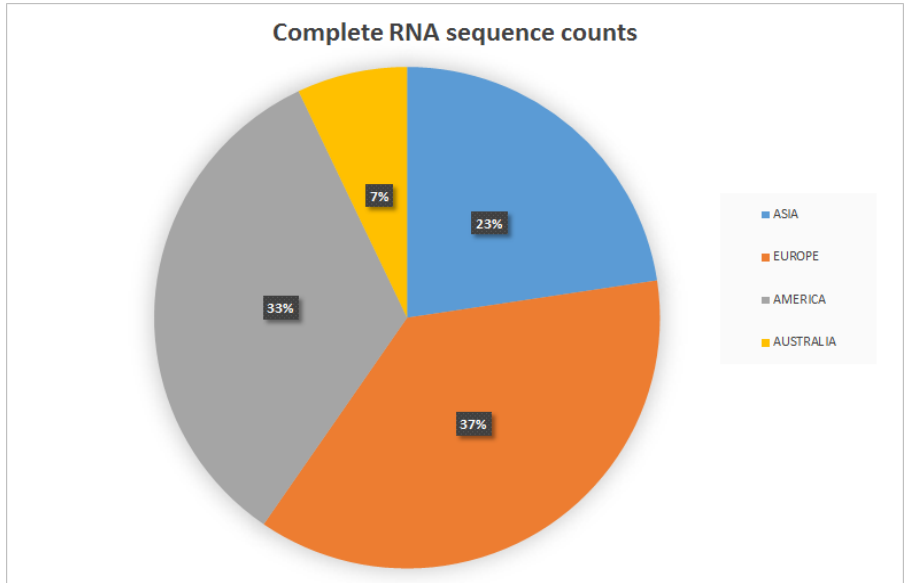

Figure 1. Distribution of complete genome sequences counts by continents.

**Creating RNA motifs with Mid-point circle algorithm**
While converting the RNA sequences of the SARS-CoV-2 virus into visual motifs, we took advantage of an analogy and expressed them as circles. An RNA sequence contains four basic nucleobases, A(denine), C(ytosine), G(uanine) and T(hymine) [12]. While expressing these nucleobases, color codes are also used. Generally, these bases are found in pairs on DNA, but these molecules are not found in pairs on RNA. This is called the open helix RNA sequence. This is one of the main differences between DNA and RNA sequences. The color codes of these nucleobases are shown in Table 1.

Table 1. Nucleobases in DNA and RNA and their color codes.[12]

| Nucleobase | Color |
| --- | --- |
| Adenine | Yellow |
| Cytosine | Blue |
| Guanine | Green |
| Thymine | Red |

The midpoint circle drawing algorithm[13] is an algorithm used to determine the points required to pixelate a circle shape. In the flow of the algorithm, to find the perimeter points of a circle, first the points in the parts divided into eight are determined, then the points in the remaining octant are determined. When determining any point (x, y) around the circle, the next pixel value is (x, y + 1) or (x-1, y + 1). We used this algorithm to fill the circle we obtained for using the RNA data that we translated into the character array. We obtained in this way with the motifs and converted the RNA sequences to 200x200 size and 3 channels into pictures. Since the gene sequences that we use in the study are not the same length, and we determined that this is the optimum dimension in our circle drawing algorithm. While creating the motifs, we used the missing and remaining pixels as white color to complete the circle. Note that a linear flat is noticeable on the right side of the circle. In terms of artificial neural networks, these pictures correspond to 200x200x3 matrices. The RNA motif that we obtained as a result of the algorithm is shown in Figure 2.

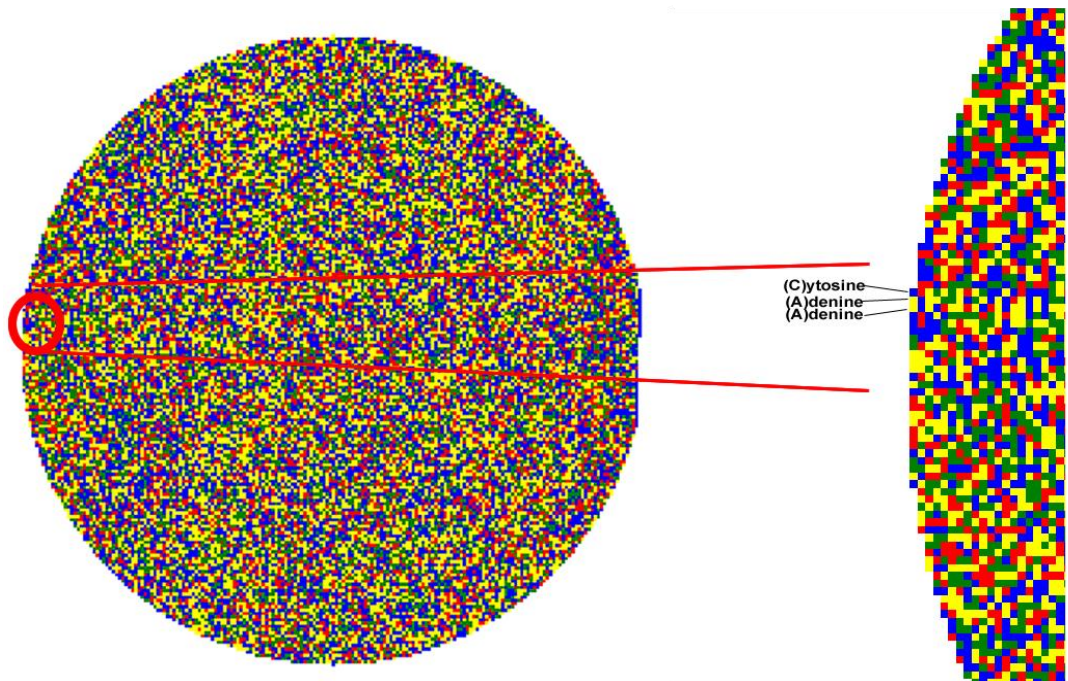

Figure 2. The motif (a) of the SARS-CoV-2 virus RNA that Accession ID EPI_ISL_425685; isolated in Scotland in March 2020, and the nucleobase equivalents (b) of the pixels within the motif.

Before obtained the images here, we used the FASTA files we obtained from datasets to convert them into motifs by separating them into RNA sequences with an application we developed with C#.

**The SUSAN Filter**

When watching the motif files we obtained from our work, and we clearly saw that some parts of it repeat. We evaluated these repetitions that we inspected may be indicative and certain patterns in the RNA structure of the virus are preserved and some parts of them are mutated. For the classification model to work correctly, identifying these patterns is important in finding mutations that differ by geographic regions. In addition, we evaluated the detection of these patterns would be beneficial in forming the phylogenetic tree structure of the newly acquired RNA sequences or in finding similarities with the previously analyzed sequences. One of the main approaches used in visual information processing such as image segmentation and pattern recognition is the edge detection method. There are many algorithms developed for edge detection today. However, low-level applications were developed at the beginning of the image processing paradigm in general and today it has been moved to higher levels. In order to detect repetitive patterns in the motif files we obtained and preferred the low-level SUSAN edge detection algorithm.

SUSAN was proposed by Smith in 1997, which was a new approach to low-level image processing. SUSAN filter consists of three main parts. These can be listed as EDGE, CORNER detections, and filtering. The mask used in SUSAN's algorithm is circular, unlike other convolution masks. The main idea of the SUSAN method is to relate each pixel of the image to a small adjacent pixel area with a brightness similar to the central pixel. The central part, called the Univalue Segment Assimilating Nucleus (USAN), is the part that conveys the most important information about the structure of the image. To summarize, the SUSAN algorithm is basically a combination of a gauss filter and a linear filter that prevents image degradation. Linear filters, in general, tend to blur more image details than most non-linear filters [14]. As a result of our applying of the SUSAN filter to the motifs we obtained, the RGB pictures were converted into grayscale. However, the dimensions of the pictures are still stored here as 200x200x3.

We also conducted tests with semantic segmentation methods to determine the patterns in the motifs we obtained during the study. But, since unsupervised segmentation methods gave different results in each motif image we obtained, we did not prefer this approach since it requires parameter adjustment for

3754 images in total. At this stage, in the SUSAN filter we used for edge detection, we found that we can mark repeating patterns in all images for the same parameters. Our results are shown in Figure 3.

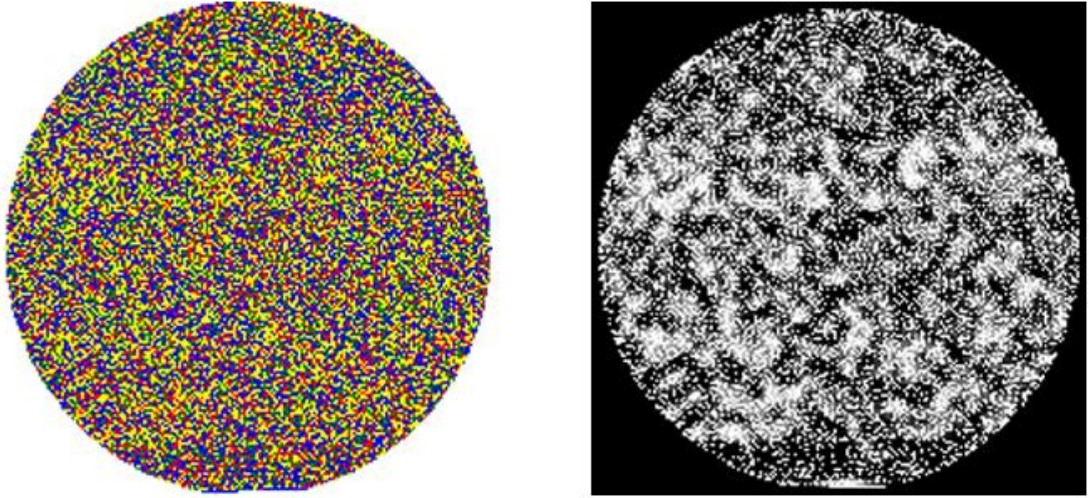

Figure 3. SUSAN filtered results for a sample motif file. The white areas in (b) show similarly repeating patterns in all motifs obtained.

**Building the model**

The creation of a CNN model in deep learning is the starting step, especially for classification success and prediction. Determining weights by building a neural network from scratch is basically a working process based on trial and error. Another criterion for the success of the models is the number and variety of data to be studied. As in this study, if the number of data is low, previously trained artificial neural network models are used. The neural networks such as Vgg16, Vgg19, Xception that have previously performed high in classification are preferred in such cases. We used a version of the DenseNet pre-trained neural network to classify the motifs we produced from this study [15]. DenseNet neural networks differ from classical convoluted networks in terms of interlayer connections. In conventional convolutional networks, there are N layers and N connections, and after applying a process combination, it connects the output of the layer to the next layer. However, there are $N(N+1)/2$ direct connections between layers in DenseNet networks. In these connections, the output feature maps of the layers are not collected, but they are combined. The connections between the layers are made in the form of feedforward. The DenseNet is divided into the DenseBlocks, where the dimensions of the feature maps remain constant within a block but the number of filters varies between them. These layers between them are called Transition Layers and are involved in sub-sampling by applying a batch normalization, 1x1 convolution, and 2x2 pooling layers. Important advantages of the DenseNet networks can be said as reducing the number of parameters and strengthen feature propagation. Apart from visual classification and clustering applications, the DenseNet neural networks are preferred in many areas such as signal processing and speech recognizing [16, 17].

If the number of objects in the dataset is low in the training of artificial neural networks, a transfer learning method is applied. This method is often preferred especially in applications such as the classification and generation of medical images. The typical way to transfer learning with deep neural networks is to fine-tune a pre-trained model on the source task using data from the target task. The transfer learning method has been preferred in computer vision applications for a long time such as domain adaptation methods. In transfer learning strategies, a modification method is used in the final layers using the ImageNet pre-trained network or its weights [18]. The learning weights of such a feed-forward network are transferred to the layers added later, resulting in higher rates of success.

CNN's architectures typically consist of different layer types, including convolution, pooling, and fully connected layers, and apply regularization for the data that has been examined. There are three main concepts in the CNN topology. These are spatial and temporal sampling, sharing weights, and receptive fields. Each convolutional layer in the CNN is made up of small kernels used to determine high-level features. The last convolutional layer in the neural network supports fully connected layers. Less connection and easy learning are provided during learning with the parameter reduction method, which is the principle of operation of the CNNs. in the first layers in the CNN, Low levels of meaningful properties such as edges, corners, textures, and lines are obtained. The kernel matrices used in this process are in the n x n dimension, and the process is carried out by advancing the number of steps called stride on the picture. The pooling layer on CNN expresses a number of fully connected layers. In this layer, high-dimensional matrices received in the previous layers are reduced to one-dimensional arrays. At the last stage, there is a stage or layer called the detector stage. In these layers, generally non-linear functions such as Softmax, ReLu, tanh are used and complex models are learned.

In this study, the weight values taken from the DenseNet-121 network were transferred to an AveragePooling layer at first. In the next step, we dropped the values from this layer to the neural network connections by 50% dropout to prevent overfitting during learning. Thanks to the dropout process, some network cells are removed from the model so that the overfitting of the neural network is prevented. At the last stage, we sent the RNA sequences to the 4 dimensional fully connected layer that will be used for the four classes we will classify. We chose the non-linear function we used in this layer as SoftMax. This function takes a vector of K real numbers as input and normalizes it to a distribution of K probabilities proportional to the exponents of the input numbers. The softmax function is expressed as

$$f(x_i) = \frac{\exp(x_i)}{\sum_j (x_j)}. \qquad (1)$$

In this way, we have obtained more than eight million parameters in the neural network model we created in this paper. However, 83.000 of these parameters are non-trainable. We used the categorical cross-entropy function for the loss function. This function is one of the common functions used for single label categorization. It indicates that only one class is applicable for each data point. The categorical cross-entropy function is defined as in Equation (2).

$$L(x, \hat{x}) = - \sum_{j=0}^{m} \sum_{i=0}^{n} (x_{ij} * \log(\hat{x}_{ij})) \qquad (2)$$

where m,n are the numbers of entries, i and j indicates iteration, and $\hat{x}$ indicates the predicted label. We preferred the optimization method and used for training as RMSProp (Root Mean Square Propagation). Optimization functions are used to determine the learning speed of the artificial neural network. We set the Learning Rate value in the optimization function to 0.001. As a result of all these values, the model and input data we obtained are shown in Figure 4.

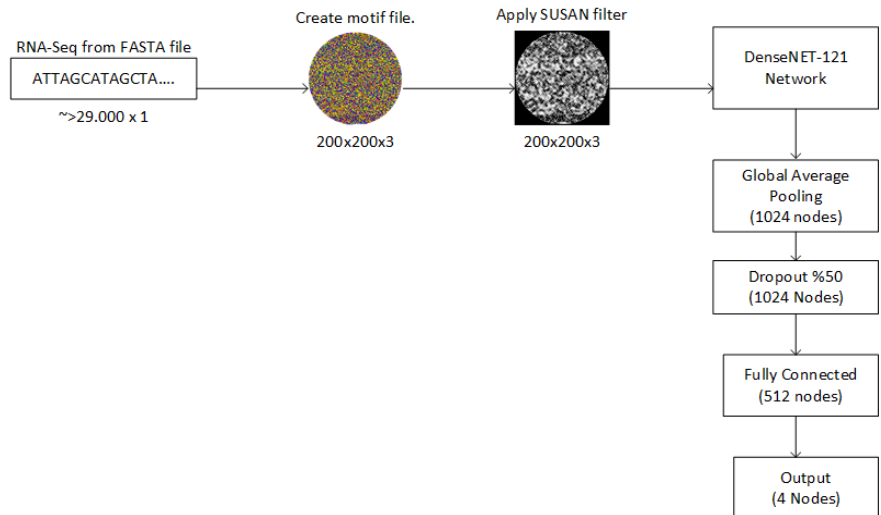

Figure 4. The stages of prepare motif files and model architecture.

We run this model on Google Colab for 75 epochs. This number is the optimum value we have obtained in our trials. Training the neural network for more epochs caused over-fitting in the data.

**Results and Discussion**

We desire to evaluate the results and obtained in this study under two headings. Overall, the success of a classification problem is evaluated by the confusion matrix and the Receiver Operating Characteristic (ROC) curve analysis. In our experiment, DenseNet-121 based one of the models we applied for the classification of four classes yielded 93% accuracy. This result is a satisfactory level of success in terms of classification. We could not achieve this level of success in the model we developed using the Vgg19 pre-trained neural network. The advantage of the DenseNet network is that although the number of parameters that can be trained by using convolution blocks increases in total, these parameters between the blocks are transferred to the next blocks, thus achieving more successful results. As a result of the filter we applied to the motif images we obtained from the RNA sequence, it is clear that this type of artificial neural network approach seems more appropriate. Our motif files have pixel sections that can be counted as noise. To generate feature maps of these pixels, they need to be subjected to a high amount of convolution and pooling. In another respect, DenseNet based models are known to give successful results in medical image classification applications today [19-21]. The visuals we obtain by applying the filter are obviously similar to medical images. The training results that we performed for the three versions of the DenseNet network together with the Vgg19 model we implemented in our first trials are shown in Table 2.

Table 2. Classification results for four different CNN models. The model we created using DenseNet-121 has optimum results.

| Model | Epochs | Region/Class | Precision | Recall | f1-score |
|---|---|---|---|---|---|
| **Vgg19** | 130 | ASIA | 0.69 | 0.74 | 0.72 |
| | | EUROPE | 0.88 | 0.85 | 0.86 |
| | | AMERICA | 0.90 | 0.83 | 0.87 |
| | | AUSTRALIA | 0.68 | 0.90 | 0.78 |
| **DenseNet-121** | 75 | ASIA | 0.88 | 0.83 | 0.86 |
| | | EUROPE | 0.93 | 0.95 | 0.94 |
| | | AMERICA | 0.98 | 0.98 | 0.98 |
| | | AUSTRALIA | 0.84 | 0.88 | 0.86 |
| **DenseNet-169** | 75 | ASIA | 0.87 | 0.87 | 0.87 |
| | | EUROPE | 0.93 | 0.95 | 0.94 |
| | | AMERICA | 0.97 | 0.98 | 0.98 |
| | | AUSTRALIA | 1.00 | 0.83 | 0.91 |
| **DenseNet-201** | 75 | ASIA | 0.88 | 0.79 | 0.83 |
| | | EUROPE | 0.90 | 0.96 | 0.93 |
| | | AMERICA | 0.98 | 0.98 | 0.98 |
| | | AUSTRALIA | 0.87 | 0.83 | 0.85 |

As can be seen in Table 2, the sensitivity of the four classes we obtained, Recall and f1 score values are quite good for the DenseNet-121 network. The training and validation performance plot of the model we trained with the DenseNet-121 network is also shown in Figure 5. Here we found the highest validation rate as 94%. However, on average, we measured the accuracy rate of our model as 93.6%.

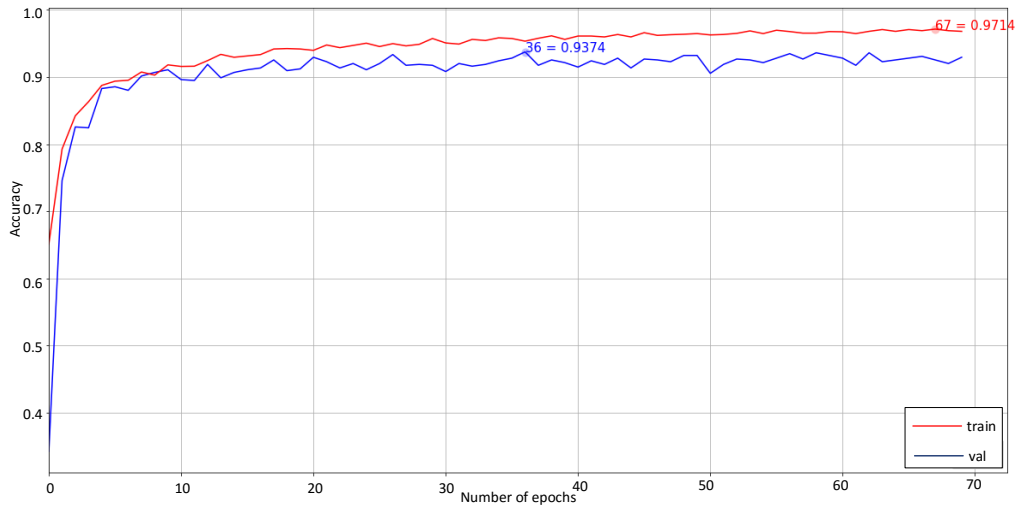

Figure 5. The plot of training and verification performance of the model.

Notably during the training of our model, although the number of data belonging to Australia is quite low, the classification success is high in the DenseNet-121 network. In such cases caused by a lack of data, the general approach is to use the augmentation method to increase model success. If we pay attention to Table 2, this model is not healthy even though the precision of the Australian class is 1.00 in the model we train in the DenseNet-169 network. This value indicates the data which is over-fitting. Moreover, although the value of Australia was such a great value in DenseNet-169, we did not get a successful result for this class in the predict we made later. In our comparison with later bioinformatics data, we chose not to evaluate the results of this model. According to the results listed in Table 2, the confusion matrix of our DenseNet-121 model, where we obtained the most optimum results, is shown in Figure 6.

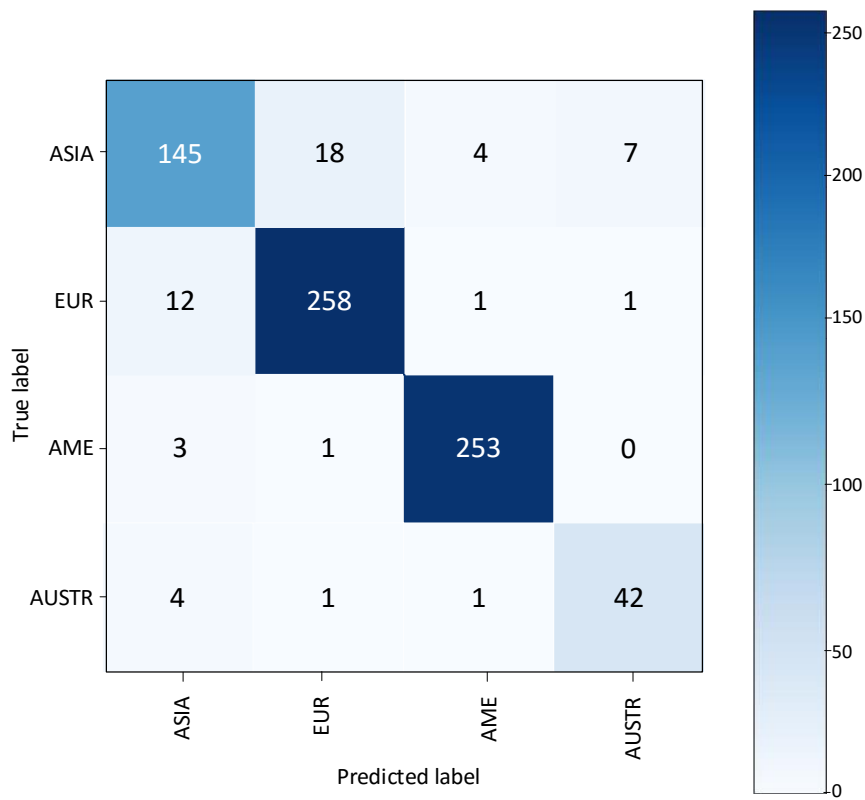

Figure 6. Confusion matrix obtained in the training of four classes.

As can be seen from the confusion matrix, a high degree of classification success is seen in our motif files. Although the incorrect values are small, considering the relationship of the data that is linked as a result of mutation, such as virus RNA, it may show their connection. Since this will be a controversial title, we consider it an issue to be emphasized. The claim that the error matrix also shows these distinctions requires a lot of research since there is no 100% geographical difference in virus mutations.

Another quantity that indicates classification success is the AUC value. In general, ROC values are defined as a probability curve in classification studies. AUC, on the other hand, expresses the degree or extent of class separability within these possibilities. The higher the AUC value, the higher the classification predictions of the model used. Here, when explaining the AUC results, values that are close to one are evaluated as the random classification of a positive sample higher than a random negative sample. The AUC values we obtained during the model training are shown in Figure 7. We classified the gene sequences that we divided into four classes into 97%, 98%, 99%, and 98% rates, respectively. These results showed us that RNA sequence information can be separated from each other by the method we use.

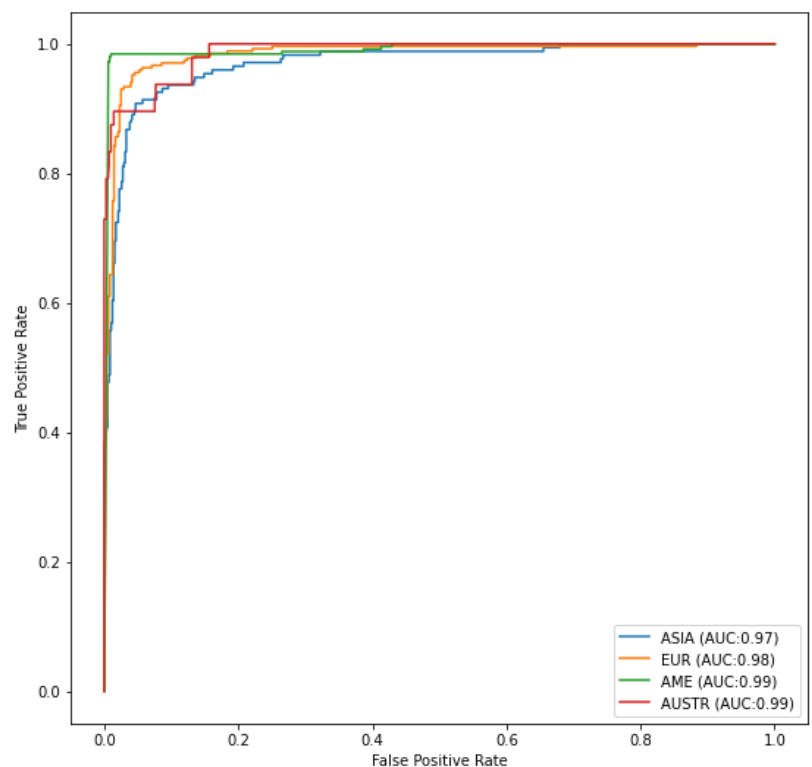

Figure 7. AUC plot of classification result for DenseNet-121

Another issue that we would like to draw attention to in our study is the difficulties in automatic pattern recognition that are still studied today. Repetitive patterns can be described as inverse of complexity. It tells us that the information obtained as a result of the observation is characterized or a name can be given to it. Examples of such patterns are fingerprints, human faces, a specific form of handwriting, or a pure signal during the speech. The two main approaches in this regard are supervised classification, which defined as discriminant examination, and unsupervised classification called clustering. In the pre-tests we conducted during this study, we tried to identify the repetitive patterns that can be easily noticed by an observer in the motif files we created with semantic segmentation. While the method we used gave good results for the object, animal, etc. encountered in daily life, it did not achieve successful results in the motif files we created. Obviously, this is because the images used to obtain weights in artificial neural networks developed for such approaches represent everyday objects. The patterns we encounter are low-level templates that can only be noticed by a human observer. In this regard, we were able to detect these repetitive unknown patterns in motif images using a low-level filter such as SUSAN. Also, another problem here was that in our dataset of approximately 3700 images, the parameters we used for semantic segmentation were only successful for one image. In this case, as many different parameters as the number of images would be required. As we have just mentioned, the image of the motif formed is a picture that contains a lot of noise. This type of pixel, normally expressed as noise, has turned into really important points in this study. In nature, a single change in any gene sequence has very different consequences. With the fixed parameters we determined in the SUSAN filter, we obtained the desired result for all images in the dataset. An example result between the two methods is shown in figure 8. For segmentation, the parameters we use here are the number of convolution layers, the number of iterations, the number of channels, and the number of labels to separate. Their values also vary according to each image in our database.

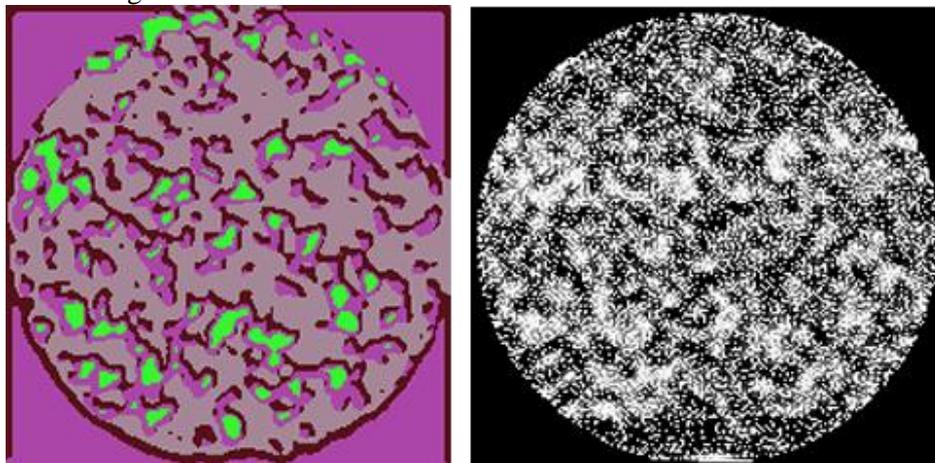

Figure 8. Semantic segmentation result (a) and SUSAN filter comparison (b) we obtained for the motif file.

Another topic that we would like to evaluate is the comparison of our classification results with a phylogenetic analysis of the isolated RNA sequences in the world. For this, we compared our classification result with the first six RNA sequences that isolated in Turkey at the GISAID EpiCoV database, which we obtained on April 30, 2020. In addition, we compared our classification results with other virus RNA data isolated in Italy, Spain, Iran, and India, where the epidemic was effective and researches were published. We have interpreted these comparisons statistically.

It is an important issue to specify the date when comparing. Because virus RNA information isolated from this date is not available in our artificial intelligence model training set. In the literature, Phylogenomic analysis methods examine the properties of 3 types of proteins to investigate the structure of these types of viruses. In particular, these classifications are important for the evolution of the S-pike protein of SARS-CoV-2, which originates from the name "Corona" and produces crown-like structures on its surface. In addition, protein regions, which are generally responsible for the formation of certain structures of viruses, have distinctions called S, V, and G clade. Making predictions and classifications for the future of a virus is possible by examining these distinctions. In studies related to the SARS-CoV-2 virus, the characteristic of S, V, and G clades reflects 69% of the virus. In addition, S-clade value indicates 72% of viruses in America and G-clade 74% of those in Europe [22]. In studies on the gene structure of the virus, which has been found is widespread episome groups according to different geographical regions such as China, Europe, and America. Although the numbers obtained from the isolation are limited, it has been shown and the classification resulting from the protein structures that we mentioned also can be grouped according to the continents [23-25]. The results obtained from these studies have shown us the visual classification study that we use here is a viable approach. However, since we did not make a comparison in the context of the V-S-G clade in this study, we only discussed our results numerically. In our comparison method, we used the proportions of the counts on four continents of the top 30 matching gene sequences of the virus analysis in the GISAID. For instance, if there are five Asian samples in 30 alignments, we accepted their numerical ratio as 16%. Unfortunately, the data we obtained from these public datasets do not have information about how many matches are in percentage. We showed our results in Table 3. Examples of the virus isolated in Turkey, not some scientific announcements, which have been mentioned links with Australia. In the results we obtained here, we observed the results confirming this connection.

Table 3. Comparison of the numerical rates of virus locations matched in the GISAID dataset and the classification rates obtained from the CNN model. Since the alignment rates were not shared in the database, we could only compare numerically.

| Nu. | Accession Id | Location | GISAID- EpiCov Our Proposal Author(s) Results(Alignment counts) | Our Proposal (Labels came from NN model) | Author(s) |
|---|---|---|---|---|---|
| 1 | EPI_ISL_424366 | Turkey | Asia 20<br>Europe 4<br>America 3<br>Australia 3 | ASIA 100% | Pavel et al |
| 2 | EPI_ISL_429864 | Turkey | Asia 8<br>Europe 18<br>America 3<br>Australia 1 | ASIA: 98.826%<br>EUR: 0.051%<br>AME: 0.001%<br>AUSTR: 1.121% | Bayrakdar et al. |
| 3 | EPI_ISL_429866 | Turkey | Asia18<br>America 12 | ASIA: 98.988%<br>EUR: 0.049%<br>AME: 0.001%<br>AUSTR: 0.962% | Bayrakdar et al. |
| 4 | EPI_ISL_429869 | Turkey | Asia 20<br>America 10 | ASIA: 98.478%<br>EUR: 0.051%<br>AME: 0.001%<br>AUSTR: 1.469% | Bayrakdar et al. |
| 5 | EPI_ISL_429870 | Turkey | Asia 10<br>Europe 16<br>America 2<br>Australia 2 | ASIA: 98.837%<br>EUR: 0.050%<br>AME: 0.001%<br>AUSTR: 1.111% | Bayrakdar et al. |
| 6 | EPI_ISL_429872 | Turkey | Asia 25<br>Australia 4 | ASIA: 98.082%<br>EUR: 0.051%<br>AME: 0.001%<br>AUSTR: 1.867% | Bayrakdar et al. |
| 7 | EPI_ISL_412973 | Italy | Asia 7<br>Europe 18<br>America 5 | ASIA: 7.254%<br>EUR: 92.219%<br>AME: 0.017%<br>AUSTR: 0.509% | Zehender et al.[26] |
| 8 | EPI_ISL_414598 | Spain | Asia 1<br>Europe28<br>America 1 | EUR: 100% | Eiez-Fuertes et al. [27] |
| 9 | EPI_ISL_413522 | India | Asia 20<br>Europe 1<br>America 9 | ASIA: 100% | Yadav et al.[28] |
| 10 | EPI_ISL_424349 | Iran | Asia 15<br>America 15 | AME 100% | Zeinali et al. |
| 11 | EPI ISL 415155 | Belgium | Asia 2<br>Europe 18<br>America 3<br>Australia7 | ASIA: 0.300%<br>EUR: 99.691%<br>AME: 0.001%<br>AUSTR: 0.008% | A. Bal et al. [29] |

In the results in Table 3, although the 7th sample is an Australian connection according to the results we find, this alignment does not appear in GISAID. However, the access numbers of EPI_ISL_428878, EPI_ISL_428915, EPI_ISL_428908, EPI_ISL_428915, EPI_ISL_428908, which are included in the top 30 alignments, have a connection with Australia. We think that we indirectly found a relationship, although it is not directly related due to the complex nature of virus RNA sequence variations. However, this claim needs to be supported by more detailed studies.

## Conclusion

When examining the overall results, we can say that the model we created is successful in terms of classification. The results showed us, thanks to the motif images that we created and the filter we used, RNA sequences can be separated according to certain classes. In half of the RNA sequences that we excluded from the CNN model training and examined the results of the comparison, we reached conclusions that could provide preliminary information about the geographic regions of the virus gene information. Here we can say that with the help of the motif of the RNA sequence we created, we have developed a method that can contribute to detecting understandable biomarkers that can be used in metagenomics studies. The succession of classification and the partial success in phylogenetic analysis is due to the need for a better filter that can be applied to the motifs we create.

Another issue to be discussed in the results that we find by obtaining a new RNA sequence of the SARS-Cov-2 virus every day, the genome alignment structure can change. Since it is still a very new field of research today, the values we find as a result of the classification can change their weight over time. Due to the random mutations of viruses, some of the RNA segments may have been randomly matched. Nevertheless, according to the information we have obtained from our motifs, certain parts of the virus RNA sequences isolated from all over the world are repeating. Although not the subject of this study, these common patterns found will contribute to the development of the drug or vaccine, against the virus.

## Acknowledgment
The GISAID and NCBI(Virus) researchers did not participate in the analysis or writing of this report. The source codes and data of the research were shared on github.com/covidcnn web address.

c